\documentclass[conference]{IEEEtran}
\IEEEoverridecommandlockouts
\usepackage{cite}
\usepackage{amsmath,amssymb,amsfonts}
\usepackage{algorithmic}
\usepackage{graphicx}
\usepackage{textcomp}
\usepackage{xcolor}
\def\BibTeX{{\rm B\kern-.05em{\sc i\kern-.025em b}\kern-.08em
    T\kern-.1667em\lower.7ex\hbox{E}\kern-.125emX}}
    
\author{Ammar Bouketta\IEEEauthorrefmark{1}\IEEEauthorrefmark{3}, Smail Niar\IEEEauthorrefmark{2}, Hamza Ouarnoughi\IEEEauthorrefmark{2}\IEEEauthorrefmark{4}, and Eva Mutuzo Brindle\IEEEauthorrefmark{3} \\
\IEEEauthorrefmark{1}Université Polytechnique Hauts-de-France, LAMIH CNRS UMR 8201, Valenciennes, France \\
\IEEEauthorrefmark{2}Université Polytechnique Hauts-de-France, LAMIH CNRS UMR 8201,  INSA Hauts-de-France, Valenciennes, France \\
\IEEEauthorrefmark{3}Alstom, Crespin, France  \\
\IEEEauthorrefmark{4}Computer Science Department, College of Computing and Informatics, University of Sharjah, Sharjah, UAE \\
Emails: 
\IEEEauthorrefmark{1}\IEEEauthorrefmark{2}\{firstname.lastname\}@uphf.fr;
\IEEEauthorrefmark{3} eva.mutuzo-brindle@alstomgroup.com
}
\begin{document}
\title{Cycle-Aware Autoencoder with Cross-Signal Consistency for Railway Door Anomaly Detection}

\maketitle
\begin{abstract}
Passenger access doors are safety-critical subsystems in railway vehicles, yet detecting abnormal door behavior in real operation is challenging because faults are rare, diverse, and often unlabeled. This paper addresses railway door condition monitoring as a cycle-level unsupervised anomaly detection problem, where each complete opening--dwell--closing cycle is treated as a single monitoring unit. We propose the Temporal Cycle-Aware Attention Autoencoder with Cross-Signal Consistency (TCAA-CS), trained exclusively on nominal cycles. It combines a dual-stream encoder that processes continuous physical measurements (position, current, voltage) and binary logical states (door-closed, door-locked) through separate 1D-CNN branches, an LSTM encoder with temporal attention pooling, and a triple hybrid anomaly score fusing reconstruction error, latent-space deviation, and phase-aware cross-signal consistency. The consistency term helps identify cases where individual signals appear plausible but their inter-signal relationships become physically or logically inconsistent. On real industrial data from a passenger train in commercial service, TCAA-CS achieves 93.8\% recall, 97.3\% precision, and a 0.5\% false-alarm rate, outperforming representative unsupervised baselines. System-level evaluation on an NVIDIA Jetson AGX Xavier supports the feasibility of real-time onboard deployment.
\end{abstract}


\section{Introduction}
Passenger access doors are safety-critical subsystems in railway vehicles. Abnormal door behavior can compromise passenger safety, reduce service availability, and increase maintenance costs. Reliable monitoring is therefore essential to support preventive maintenance and dependable railway operation. In this work, monitoring is treated as an onboard condition-monitoring function for maintenance support, rather than a certified safety-control function. Safety-critical door control remains handled by certified train systems. Monitoring passenger doors under real operating conditions is challenging. Onboard data predominantly reflect nominal behavior, while faults are rare, diverse, and generally unlabeled. Door operation is also inherently structured: each passenger access corresponds to a complete \textbf{door operation cycle}, composed of opening, dwell, and closing phases governed by coupled mechanical, electrical, and logical components. Moreover, the monitored signals are heterogeneous, combining continuous physical measurements, such as motor position, current, and supply voltage, with binary logical states, such as door-closed and door locked. Treating these signals uniformly may obscure signal-specific patterns and weaken the learned representation of normal behavior.

Abnormal behavior may not appear as an isolated signal deviation, but as an inconsistency across signals during a full door cycle. For example, a door may report a locked state while the position signal still indicates motion. Such inconsistencies are difficult to capture with methods that evaluate signals independently, rely only on reconstruction error, or operate on short windows instead of complete door cycles.

To address these challenges, we propose TCAA-CS, a cycle-level unsupervised anomaly detection framework tailored to railway door monitoring. The method combines signal-type-aware encoding, attention-based temporal representation learning, and hybrid anomaly scoring to detect both signal-level deviations and cross-signal inconsistencies.

\noindent\textbf{Contributions:} The main contributions of this work are:
\begin{itemize}
    \item We formulate railway door monitoring as a cycle-level unsupervised anomaly detection problem, where each opening--dwell--closing operation is treated as one monitoring unit.

    \item We propose TCAA-CS, a domain-tailored attention autoencoder that combines dual-stream physical/logical encoding with temporal attention pooling.

    \item We introduce a hybrid anomaly score combining reconstruction error, latent-space deviation, and phase-aware cross-signal consistency to detect both signal-level deviations and inter-signal inconsistencies.

    \item We validate TCAA-CS on real industrial railway door data, compare it with representative unsupervised baselines, and assess embedded deployment feasibility in terms of latency, memory, and energy consumption.
\end{itemize}

\section{Related Work}
\label{sec:related}

\subsection{Fault Detection in Railway Door Systems}

Railway passenger doors account for a substantial share of railway vehicle malfunctions~\cite{shimizu2023fault}, making their monitoring an important reliability and maintenance issue. Early studies mainly relied on model-based methods, such as parameter estimation from motor current signals~\cite{dassanayake2009use} and Bond Graph modeling for fault detection and isolation~\cite{cauffriez2016bond}. These approaches are interpretable, but require accurate system models and can be sensitive to modeling assumptions.

Recent work has explored data-driven diagnosis. Ham et al.~\cite{ham2019comparative} compare handcrafted features with convolutional neural networks (CNNs) applied to motor current signals, while Sun et al.~\cite{sun2020fault} use acoustic signals with empirical mode decomposition and support vector machines (SVMs). In the unsupervised setting, Shimizu et al.~\cite{shimizu2022real} combine deep autoencoders with a one-class SVM trained on nominal data, with follow-up work addressing transfer learning and generative adversarial network (GAN)-based domain adaptation across door types~\cite{shimizu2023fault}.

Despite this progress, three limitations remain. Existing methods often rely on a single signal modality, analyze samples or short windows rather than complete opening--dwell--closing cycles, and generally use a single detection criterion without explicitly assessing consistency between physical measurements and logical states.

\subsection{Anomaly Detection in Multivariate Time Series}

Unsupervised anomaly detection in multivariate time series is commonly addressed using reconstruction-based autoencoders trained on nominal data~\cite{malhotra2016lstm}. More advanced methods introduce additional criteria, such as adversarial reconstruction in UnSupervised Anomaly Detection (USAD)~\cite{audibert2020usad}, latent density estimation in the Deep Autoencoding Gaussian Mixture Model (DAGMM)~\cite{zong2018deep}, and association discrepancy in the Anomaly Transformer~\cite{xu2021anomaly}. Although effective on general benchmarks, these methods usually process fixed-length windows, treat input variables as homogeneous channels, and do not explicitly evaluate phase-wise consistency between continuous physical measurements and binary logical states.

\subsection{Positioning of This Work}

TCAA-CS addresses these gaps by aligning anomaly detection with railway door operation. It performs cycle-level detection, processes physical and logical signals through separate encoding streams, and combines reconstruction error, latent-space deviation, and phase-aware cross-signal consistency. The framework is evaluated on real industrial door data and assessed for onboard deployment feasibility in terms of latency, memory footprint, and energy consumption.

\section{Dataset Overview}
\label{sec:dataset_overview}

The dataset was collected from onboard monitoring systems of a Regio~2N passenger train operating in regular commercial service. The train, manufactured by Alstom, consists of 5 cars equipped with 16 automatic passenger access doors. In this study, we use data recorded over 30 operating days. The dataset covers multiple doors, cars, and operating days, and therefore includes natural variability induced by real commercial service.

The data predominantly reflect normal behavior, while fault events are rare and not systematically annotated. A small number of abnormal door operation cycles corresponding to real operational issues were identified through expert analysis, but they do not constitute exhaustive fault labeling. The dataset is industrial and proprietary. To improve transparency and reproducibility, we provide a precise description of the data organization, and cycle segmentation procedure.

\subsection{Cycle-Based Data Representation}
\label{subsec:cycle_representation}

\begin{figure}[htpb]
    \centering
    \includegraphics[width=1\linewidth, height=0.7\linewidth]{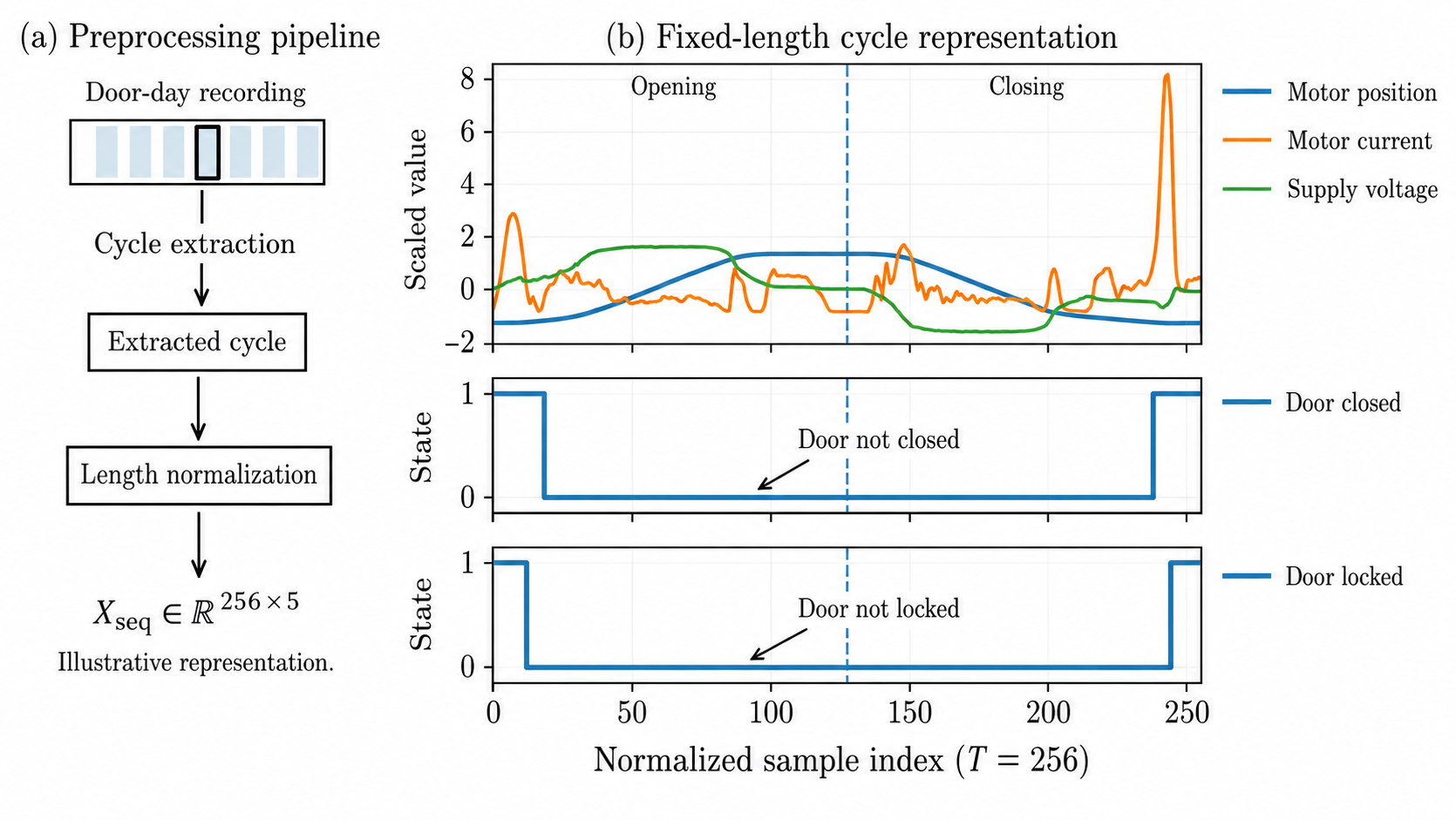}
    \caption{
    Cycle-level data representation.
    Left: segmentation of a day-long door recording into individual door operation cycles.
    Right: example of a temporally normalized cycle ($T=256$) showing continuous signals
    and logical states aligned with opening and closing phases.
    }
    \label{fig:cycle_example}
\end{figure}

Each door-day recording is segmented into \textbf{door operation cycles}. A cycle corresponds to one complete \textbf{opening--dwell--closing operation}, from the onset of door opening to the end of the subsequent closing phase. This cycle represents the natural operational unit of passenger door systems and constitutes the atomic sample used for anomaly detection. Each extracted cycle captures the mechanical, electrical, and logical behavior of the door during a single passenger access event. The five synchronized signals are acquired at a fixed sampling rate of 50~Hz and processed at the cycle level. Figure~\ref{fig:cycle_example} illustrates the transformation from raw door-day recordings to cycle-level samples used as model input.

\paragraph{Temporal resampling}
Door operation cycles have variable durations due to real operating conditions. To enable batch training and consistent cycle-level comparison, each cycle is temporally resampled to a fixed length of $T=256$ samples, while preserving the temporal ordering of the opening, dwell, and closing phases.

\paragraph{Amplitude scaling}
Continuous signals, namely door position, motor current, and supply voltage, are independently normalized to a common numerical range to ensure balanced learning across features. Binary logical state signals, namely door-closed and door-locked, are kept in their original representation $\{0,1\}$, where 0 indicates the door is open or unlocked, respectively.

After segmentation and normalization, each door operation cycle is represented as a multivariate sequence $X \in \mathbb{R}^{256 \times 5}$, capturing the synchronized temporal evolution of three continuous physical variables and two binary logical states. Each cycle is also associated with metadata, including operating day, door identifier, and original duration. These metadata are used only for traceability and result analysis, and are not provided to the model during training.

\subsection{Dataset Statistics}
\label{subsec:dataset_statistics}

Table~\ref{tab:dataset_stats} summarizes the main characteristics of the dataset after cycle segmentation and temporal normalization.

\begin{table}[htbp]
\centering
\scriptsize
\caption{Main characteristics of the railway door operation dataset.}
\label{tab:dataset_stats}
\renewcommand{\arraystretch}{1.3}
\begin{tabular}{|p{0.6cm}|p{0.6cm}|p{0.6cm}|p{0.9cm}|p{1.4cm}|p{1.1cm}|p{0.6cm}|}
\hline
\textbf{Days} & \textbf{Cars} & \textbf{Doors} &
\textbf{Cycles} & \textbf{Features ($F$)} & \textbf{Cycle length ($T$)} & \textbf{Freq. (Hz)} \\
\hline
30 & 5 & 16 & 4\,960 & 5 & 256 & 50 \\
\hline
\end{tabular}
\end{table}

\section{Methodology}
\label{sec:methodology}

This section presents the proposed anomaly detection framework for railway passenger door monitoring. Following the cycle-based representation introduced in Section~\ref{sec:dataset_overview}, detection is performed at the cycle level: each complete door operation constitutes one monitoring instance and produces one anomaly score and one decision.

\subsection{Problem Setting}
\label{subsec:problem_setting}

Each door operation cycle is represented as a fixed-length multivariate time series
\begin{equation}
X \in \mathbb{R}^{T \times F},
\label{eq:input}
\end{equation}
where $T=256$ is the normalized cycle length and $F=5$ is the number of synchronized signals, including three continuous physical signals and two binary logical states.

Fault labels are generally unavailable in real operation, and abnormal events are rare. The model is therefore trained exclusively on nominal cycles. Given an unseen test cycle $X_j$, the objective is to compute a scalar anomaly score $S(X_j)$ that measures its deviation from nominal behavior. This score is compared with a threshold $\tau$ calibrated on nominal validation data, and the cycle is declared anomalous when $S(X_j)>\tau$.

\subsection{Model Overview}
\label{subsec:model_overview}

\begin{figure}[htbp]
    \centering
    \includegraphics[width=\linewidth]{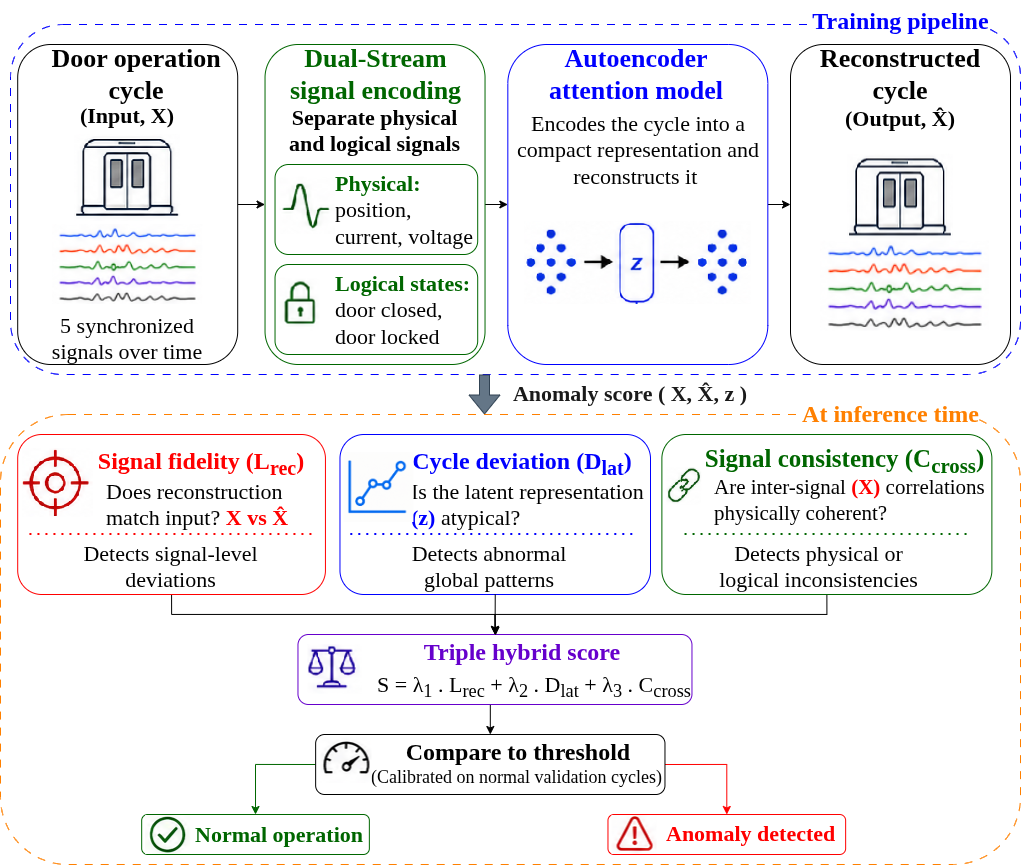}
    \caption{High-level overview of the TCAA-CS framework. Nominal cycles are used to train the reconstruction model, while inference combines signal fidelity, latent-space deviation, and cross-signal consistency into one anomaly decision per cycle.}
    \label{fig:overview}
\end{figure}

Figure~\ref{fig:overview} provides a high-level overview of the proposed TCAA-CS framework. During training, each door operation cycle is processed by a \textbf{dual-stream signal encoding} stage, which separates continuous physical measurements from binary logical states. The resulting representation is then passed to an \textbf{attention-based autoencoder}, which compresses the cycle into a compact latent representation and reconstructs the original sequence. The model is trained on nominal cycles only by minimizing reconstruction error.

At inference time, TCAA-CS assigns one anomaly score to each complete door operation cycle. This score combines three complementary criteria: \textbf{signal fidelity}, measured by reconstruction error; \textbf{latent-space deviation}, which quantifies how far the cycle representation deviates from nominal latent behavior; and \textbf{cross-signal consistency}, which evaluates whether phase-wise inter-signal relationships remain coherent with nominal operation. These criteria are fused into a single triple hybrid anomaly score and compared against a calibrated threshold.

\begin{figure*}[t]
    \centering
    \includegraphics[width=\linewidth]{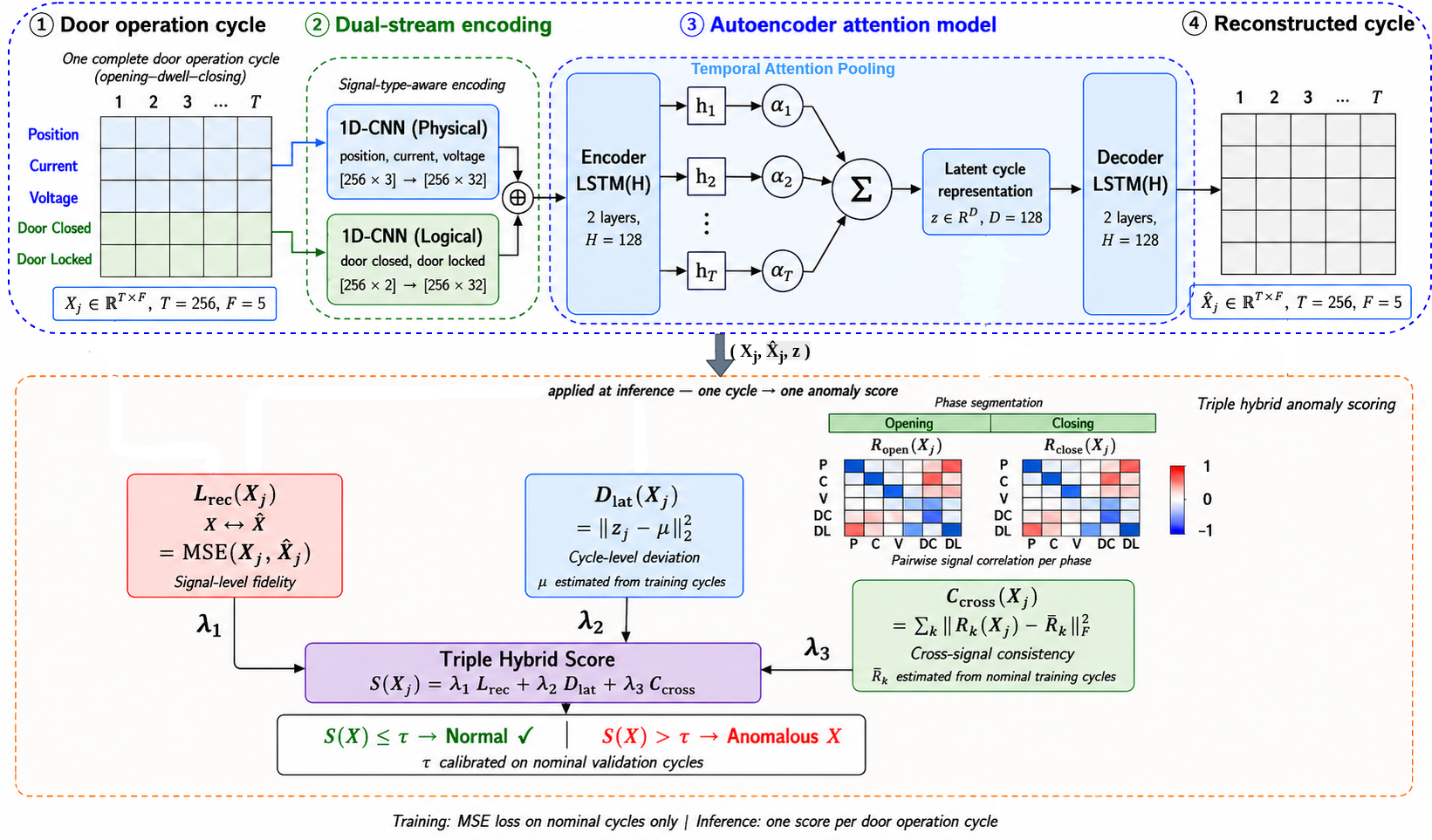}
    \caption{Detailed architecture of TCAA-CS for cycle-level railway door anomaly detection. The upper part shows dual-stream signal encoding, attention-based reconstruction, and latent representation learning. The lower part shows the triple hybrid anomaly scoring mechanism.}
    \label{fig:architecture}
\end{figure*}

Figure~\ref{fig:architecture} presents the detailed TCAA-CS architecture. The upper part shows dual-stream encoding, temporal modeling with LSTM and attention pooling, and reconstruction through an LSTM decoder. The lower part shows the triple hybrid scoring mechanism derived from the input cycle $X$, the reconstructed cycle $\hat{X}$, and the latent representation $z$.

\subsection{Dual-Stream Signal Encoding}
\label{subsec:dual_stream}

The five input signals have different statistical properties. Motor position, motor current, and supply voltage are continuous measurements that describe mechanical and electrical dynamics, whereas door-closed and door-locked are binary logical states. Processing all signals through the same initial representation treats them as homogeneous channels and may obscure signal-type-specific patterns.

To preserve this distinction, TCAA-CS separates the input cycle into physical and logical streams, denoted by $X^{\text{phys}}\in\mathbb{R}^{T\times3}$ and $X^{\text{log}}\in\mathbb{R}^{T\times2}$. Each stream is processed by a dedicated temporal convolutional branch:
\[
H^{\text{phys}}=\mathrm{CNN}_{\text{phys}}(X^{\text{phys}}), \qquad
H^{\text{log}}=\mathrm{CNN}_{\text{log}}(X^{\text{log}}),
\]
where both outputs are in $\mathbb{R}^{T\times d_c}$ and $d_c=32$. The physical branch learns local position, current, and voltage patterns, while the logical branch learns state-transition patterns. The two encoded streams are concatenated as
\begin{equation}
H^{\text{in}}=
[H^{\text{phys}} \,\|\, H^{\text{log}}]
\in \mathbb{R}^{T \times 2d_c}.
\label{eq:dual_stream}
\end{equation}
This produces a unified $T \times 64$ feature sequence while preserving the distinction between physical and logical behavior before joint temporal modeling.

\subsection{LSTM Encoder with Temporal Attention Pooling}
\label{subsec:encoder_attention}

The fused feature sequence $H^{\text{in}}$ is processed by a two-layer LSTM encoder with hidden dimension $d=128$, producing hidden states $(h_1,\ldots,h_T)$, where each $h_t\in\mathbb{R}^{d}$ represents the temporal context at timestep $t$.

A standard LSTM autoencoder often represents the full sequence using only the final hidden state $h_T$. This can be limiting for railway door cycles because informative events may occur at different phases of the operation, such as motion onset, phase transitions, closing impact, or lock engagement. TCAA-CS instead constructs the cycle-level representation using additive temporal attention pooling. A relevance score is computed for each hidden state:
\begin{equation}
e_t = v^\top \tanh(W h_t + b),
\end{equation}
where $W$, $v$, and $b$ are learnable parameters. The scores are normalized with a softmax function:
\begin{equation}
\alpha_t =
\frac{\exp(e_t)}
{\sum_{k=1}^{T} \exp(e_k)}.
\end{equation}
The latent representation of the full cycle is then obtained as
\begin{equation}
z =
\sum_{t=1}^{T} \alpha_t h_t
\in \mathbb{R}^{d}.
\end{equation}

The attention weights $\alpha_t$ allow the model to emphasize informative parts of the opening--dwell--closing operation while reducing the influence of less informative segments.

\subsection{Decoder and Training Objective}
\label{subsec:decoder}

The latent representation $z$ is passed to a symmetric two-layer LSTM decoder, which reconstructs the complete multivariate sequence:
\[
\hat{X} = \mathrm{Decoder}(z) \in \mathbb{R}^{T \times F}.
\]

The model is trained using nominal cycles only by minimizing the mean squared reconstruction error:
\begin{equation}
\mathcal{L}_{\text{train}}
=
\frac{1}{N}
\sum_{i=1}^{N}
\mathrm{MSE}(X_i,\hat{X}_i),
\end{equation}

where $N$ is the number of nominal training cycles, $X_i$ denotes the $i$-th nominal training cycle, and $\hat{X}_i$ is its reconstruction. The MSE is computed over all timesteps and signal dimensions. After training, deviations from learned nominal behavior are quantified through the triple hybrid anomaly score.

\subsection{Triple Hybrid Anomaly Scoring}
\label{subsec:triple_scoring}

At inference time, TCAA-CS evaluates each test cycle $X_j$ using three complementary anomaly indicators. Reconstruction error captures signal-level deviations, latent-space deviation captures global cycle-level abnormality, and cross-signal consistency captures violations of expected relationships between heterogeneous signals.

\subsubsection{Reconstruction Error}

The first component measures how accurately the model reconstructs the observed test cycle:
\begin{equation}
\mathcal{L}_{\text{rec}}(X_j)
=
\mathrm{MSE}(X_j,\hat{X}_j).
\label{eq:rec_score}
\end{equation}
This score captures deviations that alter the expected shape or amplitude of one or more signals, such as abnormal current peaks or distorted motion profiles.

\subsubsection{Latent-Space Deviation (Cycle-Level Abnormality)}

The second component measures whether the latent representation of a test cycle is consistent with nominal cycle behavior. The nominal latent center is computed from the training cycles:
\begin{equation}
\mu =
\frac{1}{N}
\sum_{i=1}^{N}
z_i.
\end{equation}
Here, $z_i$ is the latent representation of the $i$-th nominal training cycle $X_i$. For a test cycle $X_j$ with latent representation $z_j$, the latent-space deviation is defined as
\begin{equation}
\mathcal{D}_{\text{lat}}(X_j)
=
\|z_j-\mu\|_2^2.
\end{equation}
This term captures cycles whose global temporal structure differs from nominal behavior, including unusual timing, abnormal phase durations, or atypical overall dynamics.

\subsubsection{Phase-Aware Cross-Signal Consistency}

The third component evaluates whether relationships between door signals remain consistent with nominal operation. During normal opening and closing, physical measurements and logical states evolve in coordinated ways. A cycle may therefore be abnormal even when individual signals appear locally plausible, if their inter-signal relationships become physically or logically inconsistent.

Each cycle is divided into opening and closing phases based on the position trajectory after temporal normalization. For each phase $k\in\{\mathrm{open},\mathrm{close}\}$, we compute a pairwise Pearson correlation matrix $R_k(X_j)\in\mathbb{R}^{F\times F}$, where each entry measures how two signals co-vary during that phase. For each phase, a nominal reference matrix is estimated from the nominal training cycles as
\begin{equation}
\bar{R}_k =
\frac{1}{N}
\sum_{i=1}^{N}
R_k(X_i).
\end{equation}
The cross-signal consistency score is then defined as
\begin{equation}
\mathcal{C}_{\text{cross}}(X_j)
=
\sum_{k \in \{\mathrm{open},\,\mathrm{close}\}}
\left\|
R_k(X_j)-\bar{R}_k
\right\|_F^2 .
\label{eq:cross_signal_score}
\end{equation}
A high value indicates that the relationships between signals during opening or closing deviate from nominal behavior. This term is particularly useful for detecting physical or logical inconsistencies, such as the door-locked signal being active while the position signal still indicates motion.

\subsubsection{Score Normalization and Fusion}

The three anomaly components have different numerical scales. Before fusion, each component is standardized using statistics computed on nominal validation data:
\begin{equation}
\tilde{Q}(X_j)
=
\frac{Q(X_j)-\mu_Q}{\sigma_Q+\epsilon},
\label{eq:score_norm}
\end{equation}
where $Q$ denotes any score component, $\mu_Q$ and $\sigma_Q$ are computed on the nominal validation set, and $\epsilon$ is a small constant for numerical stability.

The final anomaly score is defined as
\begin{equation}
S_{\text{TCAA-CS}}(X_j)
=
\lambda_1 \tilde{\mathcal{L}}_{\text{rec}}(X_j)
+
\lambda_2 \tilde{\mathcal{D}}_{\text{lat}}(X_j)
+
\lambda_3 \tilde{\mathcal{C}}_{\text{cross}}(X_j),
\label{eq:final_score}
\end{equation}
where $\lambda_1,\lambda_2,\lambda_3\geq0$ and $\lambda_1+\lambda_2+\lambda_3=1$. The weights control the relative contribution of each anomaly component and are selected through the sensitivity analysis reported in Section~\ref{sec:ablation}.

\subsection{Decision Rule}
\label{subsec:decision}

The decision threshold $\tau$ is calibrated exclusively on nominal validation data as a high percentile of the nominal validation score distribution:
\[
\tau =
\mathrm{Percentile}_{p}
\left(
\{S_{\text{TCAA-CS}}(X_m)\}_{X_m \in \mathcal{D}_{\text{val}}}
\right),
\]
where $X_m$ denotes a nominal validation cycle and $p$ controls the tolerated false-alarm level. A test cycle $X_j$ is classified as
\begin{equation}
\hat{y}_j =
\begin{cases}
1, & \text{if } S_{\text{TCAA-CS}}(X_j) > \tau,\\
0, & \text{otherwise},
\end{cases}
\label{eq:decision}
\end{equation}
where $\hat{y}_j=1$ denotes an anomalous cycle and $\hat{y}_j=0$ denotes a nominal cycle. This formulation produces one decision per complete door operation cycle and requires only nominal data for training and calibration.

\section{Experimental Evaluation}
\label{sec:evaluation}

This section evaluates the proposed TCAA-CS framework on real-world railway door operation data. All experiments follow a fixed and reproducible protocol. We first describe the evaluation setup, then report detection results against representative unsupervised baselines. We then present ablation studies to justify the main design choices, followed by system-level deployment experiments and qualitative analysis of real abnormal cycles.

\subsection{Experimental Protocol}
\label{subsec:experimental_protocol}

\subsubsection{Dataset Split}

The dataset contains 4\,960 door operation cycles collected over 30 operating days from 16 doors across 5 cars. To evaluate generalization across operating conditions, a day-wise split is adopted: all cycles from a given day are assigned to the same subset, and no day appears in more than one split. Table~\ref{tab:split_summary} reports the resulting split and the final test-set composition after test-time anomaly injection.

\begin{table}[htbp]
\caption{Day-wise train/validation/test split and final test-set composition used for evaluation.}
\begin{center}
\begin{tabular}{|l|c|c|c|c|}
\hline
 & \textbf{Days} & \textbf{Cycles} & \textbf{Nominal} & \textbf{Anomalous} \\
\hline
Training   & 20 & 3\,179 & 3\,179 & 0 \\
\hline
Validation & 4  & 620    & 620    & 0 \\
\hline
Test       & 6  & 1\,161 & 969    & 192 \\
\hline
\end{tabular}
\label{tab:split_summary}
\end{center}
\end{table}

Expert-identified abnormal cycles are excluded from the training and validation subsets and retained only for test evaluation. Consequently, the training and validation sets contain nominal cycles only. The validation set is used for early stopping, score normalization, and threshold calibration. Anomalous cycles are used exclusively in the test set.

\subsubsection{Test-Time Anomaly Construction}
\label{subsec:anomaly_injection}

Real abnormal door cycles are present but remain limited compared with nominal operation. Over the 30 operating days, 12 abnormal cycles were identified and confirmed through expert analysis. Since these events alone are insufficient for class-level quantitative evaluation, additional anomalies are introduced through controlled injection at test time only.

The injection mechanisms reproduce fault patterns derived from documented maintenance records and expert observations of real operational failures. Each anomaly class corresponds to a known failure mode encountered in railway door systems. The injection parameters and resulting signal profiles were validated by domain engineers to ensure physical and operational plausibility.
 Table~\ref{tab:anomaly_classes} summarizes the injected anomaly classes.

\begin{table}[htbp]
\caption{Abnormal cycle classes used for test-time evaluation.}
\begin{center}
\begin{tabular}{|p{0.6cm}|p{1.1cm}|p{5.9cm}|}
\hline
\textbf{Class} & \textbf{Type}  & \textbf{Description} \\
\hline
A & Energy / Load  & Increased or biased motor current and/or voltage during opening or closing, simulating friction. \\
\hline
B & Temporal/ Transient  & Local stretching or compression of motion phases, short spikes, dropouts, or sensor noise. \\
\hline
C & Logical & Inconsistent signals, e.g., door locked active while the door is moving. \\
\hline
\end{tabular}
\label{tab:anomaly_classes}
\end{center}
\end{table}

From the nominal portion of the test set, 180 cycles are selected using a fixed random seed, with 60 cycles assigned to each injected anomaly class. These cycles are replaced by perturbed versions, each containing exactly one injected anomaly. Together with the 12 real expert-confirmed abnormal cycles, the final test set contains 192 anomalous cycles and 969 nominal cycles. Injected anomalies are never used during training, validation, early stopping, score normalization, or threshold calibration.

\subsubsection{Evaluation Metrics}

Performance is assessed using recall, precision, F1-score, and false-alarm rate (FAR). Recall is the primary metric because missed anomalies correspond to abnormal door operations that remain undetected. FAR is computed on nominal test cycles only and reflects the proportion of nominal cycles incorrectly flagged as anomalous, which directly affects maintenance workload.

\subsubsection{Compared Methods}

The proposed method is compared against unsupervised baselines representing complementary modeling paradigms:

\begin{itemize}
    \item \textbf{LSTM-AE}: recurrent autoencoder using the final hidden state as cycle representation, with reconstruction error as anomaly score~\cite{malhotra2016lstm}.
    \item \textbf{GRU-AE}: recurrent autoencoder using GRU units instead of LSTM units~\cite{gong2022autoencoder}.
    \item \textbf{TCN-AE}: temporal convolutional autoencoder~\cite{thill2021temporal}.
    \item \textbf{OC-SVM}: one-class SVM trained on handcrafted cycle-level features~\cite{scholkopf2001estimating}.
    \item \textbf{Anomaly Transformer}: transformer-based anomaly detection model using association discrepancy~\cite{xu2021anomaly}. For fair comparison, it is trained and evaluated on the same fixed-length cycle representation, and timestep-level anomaly scores are aggregated to produce one score per cycle.
\end{itemize}

All methods are trained on nominal data only, use comparable training budgets with early stopping, and produce one anomaly score per cycle. Thresholds are calibrated using the same percentile-based rule on nominal validation data.

\subsection{TCAA-CS Configuration}
\label{sec:tcaa_config}

\paragraph{Architecture}
The dual-stream encoder uses two 1D-CNN branches, each with two convolutional layers, kernel size 5, output dimension $d_c=32$, ReLU activation, and batch normalization. The LSTM encoder and decoder each have two layers with hidden dimension $d=128$. Dropout of 0.15 is applied between recurrent layers.

\paragraph{Training}
The model is trained by minimizing reconstruction error using AdamW, with batch size 64 and a maximum of 120 epochs. Early stopping is applied based on nominal validation loss.

\paragraph{Scoring weights}
The triple hybrid score weights are fixed to $\lambda_1=0.6$, $\lambda_2=0.1$, and $\lambda_3=0.3$ based on the sensitivity analysis reported in Section~V-D. The decision threshold is set at the $p=95$-th percentile of the nominal validation score distribution. This percentile was selected to favor high recall while keeping the false-alarm rate low; it can be adjusted depending on the maintenance operator's tolerance to false alarms.

\subsection{Main Detection Results}
\label{subsec:main_results}

Table~\ref{tab:main_results} summarizes detection performance on the test set, which contains 192 anomalous cycles, including 12 real expert-confirmed anomalies and 180 injected anomalies, together with 969 nominal cycles.

\begin{table}[htbp]
\caption{Cycle-level anomaly detection results. FAR is computed on nominal test cycles only.}
\begin{center}
\begin{tabular}{|p{1.7cm}|p{0.3cm}|p{0.3cm}|p{0.3cm}|p{0.3cm}|p{0.6cm}|p{0.5cm}|p{0.5cm}|p{0.5cm}|}
\hline
\textbf{Method} & $TP$ & $FN$ & $FP$ & $TN$ & \textbf{Recall} & \textbf{Prec.} & \textbf{F1} & \textbf{FAR} \\
\hline
LSTM-AE~\cite{malhotra2016lstm}        & 145 & 47 & 22 & 947 & 0.755 & 0.868 & 0.808 & 0.023 \\
\hline
GRU-AE~\cite{gong2022autoencoder}         & 163 & 29 & 20 & 949 & 0.849 & 0.891 & 0.870 & 0.021 \\
\hline
TCN-AE~\cite{thill2021temporal}         & 136 & 56 & 16 & 953 & 0.708 & 0.895 & 0.791 & 0.017 \\
\hline
OC-SVM~\cite{scholkopf2001estimating}         & 120 & 72 & 30 & 939 & 0.625 & 0.800 & 0.702 & 0.031 \\
\hline
 Transformer~\cite{xu2021anomaly}   & 158 & 34 & 18 & 951 & 0.823 & 0.898 & 0.859 & 0.019 \\
\hline
\textbf{TCAA-CS} & \textbf{180} & \textbf{12} & \textbf{5} & \textbf{964} & \textbf{0.938} & \textbf{0.973} & \textbf{0.955} & \textbf{0.005} \\
\hline
\end{tabular}
\label{tab:main_results}
\end{center}
\end{table}

The baseline methods achieve competitive performance, confirming that standard unsupervised sequence models capture a substantial part of abnormal door-cycle behavior. GRU-AE provides the strongest baseline, with 84.9\% recall and an F1-score of 0.870, followed by the Anomaly Transformer with 82.3\% recall and an F1-score of 0.859. TCN-AE shows high precision but lower recall, suggesting a more conservative detection behavior, while OC-SVM remains less competitive because it relies on handcrafted cycle-level features rather than learned temporal representations.

TCAA-CS achieves the best overall performance, with 93.8\% recall, 97.3\% precision, an F1-score of 0.955, and a false-alarm rate of 0.5\%. Compared with the strongest baseline, GRU-AE, TCAA-CS improves recall by 8.9 percentage points and F1-score by 8.5 percentage points under the same threshold calibration protocol. This improvement suggests that dual-stream signal encoding and triple hybrid scoring capture anomaly patterns that are not fully addressed by reconstruction-based recurrent baselines.

Out of 192 anomalous cycles, TCAA-CS correctly detects 180 and misses 12. Notably, all 12 expert-confirmed real operational anomalies are detected, whereas none of the baseline methods detects all of them. The 12 missed detections correspond to low-magnitude controlled perturbations with amplitudes close to nominal variability. The method produces 5 false alarms on 969 nominal cycles, mainly arising from rare but valid operational events such as passenger-triggered reopening commands or atypical but non-faulty cycle timing.

\subsection{Ablation Study}
\label{sec:ablation}

To justify the main design choices, we perform a systematic ablation study. Starting from the full TCAA-CS model, we individually remove key components and measure the impact on detection performance. When a scoring component is removed, the remaining score weights are renormalized to sum to one, and the decision threshold is recalibrated on nominal validation scores using the same percentile rule.

\subsubsection{Component Ablation}

Table~\ref{tab:ablation_components} reports the effect of removing each key component from the full model.

\begin{table}[htbp]
\caption{Component ablation. Each row removes one component from the full TCAA-CS model; ``w/o'' denotes ``without''.}
\begin{center}
\begin{tabular}{|l|c|c|c|c|}
\hline
\textbf{Configuration} & \textbf{Recall} & \textbf{Prec.} & \textbf{F1} & \textbf{FAR} \\
\hline
Full TCAA-CS & \textbf{0.938} & \textbf{0.973} & \textbf{0.955} & \textbf{0.005} \\
\hline
w/o dual-stream & 0.906 & 0.946 & 0.925 & 0.012 \\
\hline
w/o attention ($h_T$ only) & 0.880 & 0.939 & 0.909 & 0.014 \\
\hline
w/o $\mathcal{C}_{\text{cross}}$ & 0.891 & 0.950 & 0.920 & 0.011 \\
\hline
w/o $\mathcal{D}_{\text{lat}}$ & 0.917 & 0.941 & 0.929 & 0.014 \\
\hline
$\mathcal{L}_{\text{rec}}$ only & 0.781 & 0.926 & 0.847 & 0.015 \\
\hline
\end{tabular}
\label{tab:ablation_components}
\end{center}
\end{table}

Removing any single component degrades performance, confirming that each contributes to overall detection. The largest drop occurs when temporal attention is removed, with recall decreasing from 93.8\% to 88.0\%. This indicates that the attention-based cycle representation is important for capturing informative phases of the door operation. Removing the cross-signal consistency term reduces recall to 89.1\%, with the loss concentrated on logical anomalies where individual signals appear plausible but their relationships are inconsistent. Removing dual-stream encoding reduces recall to 90.6\%, confirming that separate processing of physical and logical signals improves the learned representation. Using reconstruction error alone reduces recall to 78.1\%, demonstrating the necessity of combining complementary anomaly indicators.

\subsubsection{Score Weight Sensitivity}
We evaluate the sensitivity of the triple hybrid score to the choice of weights using a grid with increments of 0.1 under the constraint $\lambda_1+\lambda_2+\lambda_3=1$. The selected configuration $\lambda_1=0.6$, $\lambda_2=0.1$, and $\lambda_3=0.3$ achieved the best F1-score, with recall 0.938, precision 0.973, F1-score 0.955, and FAR 0.005. Neighboring configurations remained competitive, such as $(0.5,0.2,0.3)$ with F1-score 0.952 and $(0.7,0.1,0.2)$ with F1-score 0.950, indicating limited sensitivity to small weight variations. In contrast, single-component settings were less effective, with reconstruction only $(1,0,0)$ reaching F1-score 0.847, latent deviation only $(0,1,0)$ reaching 0.474, and cross-signal consistency only $(0,0,1)$ reaching 0.456. These results support the use of a reconstruction-dominant but complementary hybrid score.

\subsection{System-Level Evaluation}
\label{subsec:system_eval}

Beyond detection accuracy, onboard deployment requires low inference latency, limited memory usage, and feasible energy consumption. We therefore evaluate the computational cost of the final TCAA-CS model during cycle-by-cycle inference. Each measurement includes the complete inference pipeline for one completed door cycle: forward pass, anomaly score computation, and decision rule. Latency is measured after a warm-up phase and averaged over repeated inference runs to reduce initialization effects.

Table~\ref{tab:system_results} reports the results on three execution environments: an Intel i7-8850H CPU, an NVIDIA Quadro P2000 desktop GPU, and an NVIDIA Jetson AGX Xavier embedded platform. In the table, \emph{mean lat.} denotes the average inference latency per cycle, \emph{p95 lat.} denotes the 95th percentile latency, \emph{max lat.} denotes the maximum observed latency, and \emph{GPU mem.} denotes peak GPU memory usage during inference.

\begin{table}[htbp]
\caption{System-level performance of TCAA-CS under cycle-by-cycle inference.}
\begin{center}
\begin{tabular}{|l|p{1cm}|p{0.7cm}|p{0.7cm}|p{0.7cm}|p{0.9cm}|}
\hline
\textbf{Platform} & \textbf{Mean lat. (ms)} & \textbf{p95 lat. (ms)}& \textbf{Max lat. (ms)}  & \textbf{GPU mem. (MB)} & \textbf{Energy} \\
\hline
Intel i7-8850H CPU & 8.52 & 10.41 & 14.8 &-- & -- \\
\hline
Quadro P2000 GPU & 5.21 & 6.08 & 8.23 & 32.7 & -- \\
\hline
Jetson AGX Xavier & 4.98 & 6.52 & 8.92 & 32.7 & 8--12\,W \\
\hline
\end{tabular}
\label{tab:system_results}
\end{center}
\end{table}

On the embedded target, Jetson AGX Xavier, TCAA-CS achieves a mean inference latency of 4.98\,ms per cycle, with a p95 latency of 6.52\,ms and maximum latency of 8.92\,ms. Since a complete door operation typically lasts several seconds, this latency is negligible compared with the physical duration of the monitored process. The small gap between mean and p95 latency also indicates stable execution under cycle-by-cycle inference.  Across GPU-based platforms, peak memory usage remains below 34\,MB, which is consistent with the compact model size of approximately 560k trainable parameters and leaves margin for a shared onboard monitoring unit to supervise multiple doors or run additional onboard modules. Energy consumption on the Jetson platform is approximately 8--12\,W under inference load. These results support the feasibility of real-time onboard deployment for cycle-level railway door monitoring.

\subsection{Qualitative Analysis on Real Anomalies}
\label{subsec:qualitative_analysis}

\begin{figure}[htpb]
\centering
\includegraphics[width=\linewidth, height=0.8\linewidth]{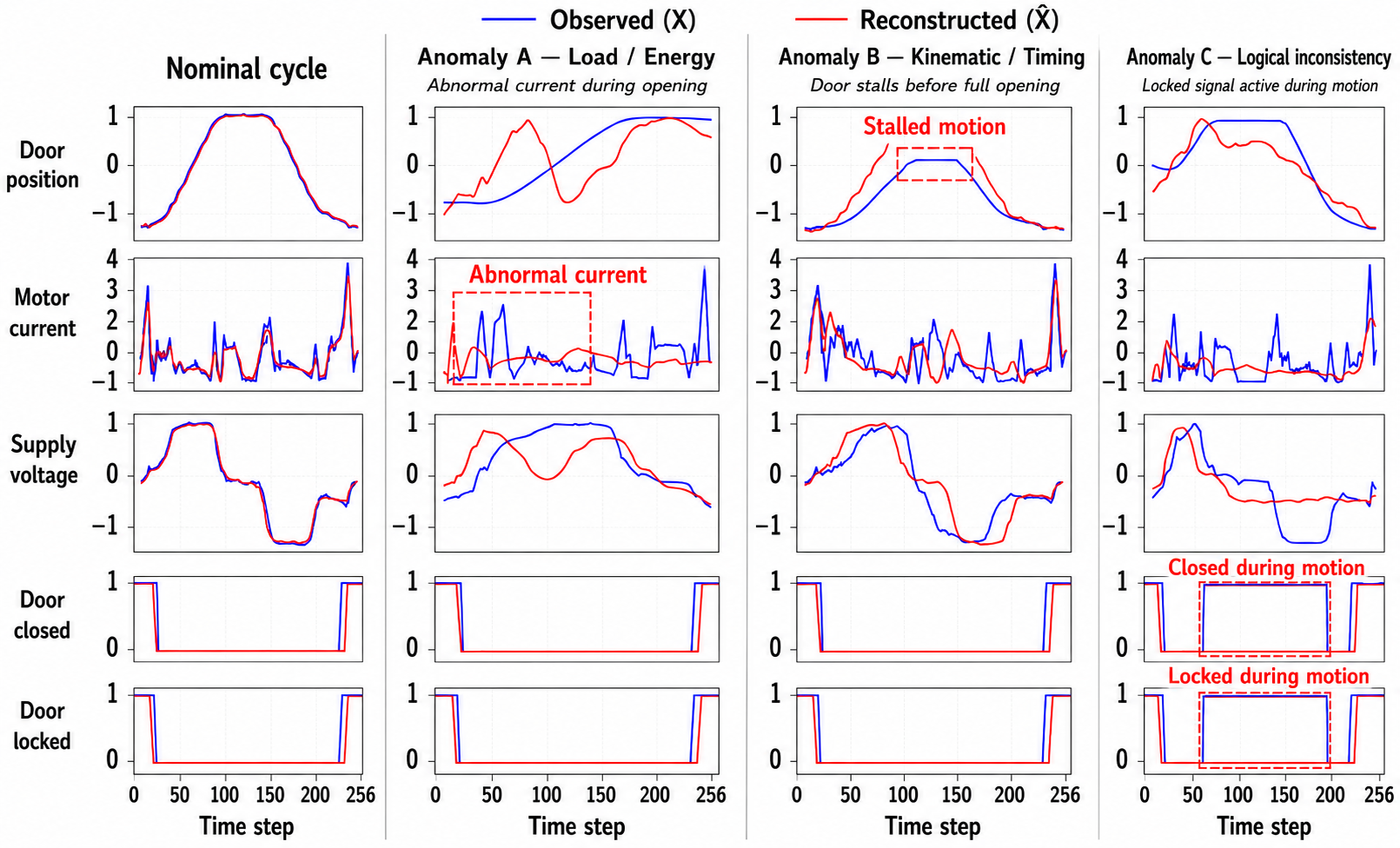}
\caption{Representative door cycles: one nominal cycle and three anomalous cycles illustrating load-related, kinematic, and logical inconsistency fault mechanisms. Blue: observed signals. Red: TCAA-CS reconstruction. Red dashed boxes: anomaly regions.}
\label{fig:qualitative_cycles}
\end{figure}

Figure~\ref{fig:qualitative_cycles} presents one nominal cycle and three anomalous cycles illustrating distinct fault mechanisms. Each column shows a complete door cycle with observed signals in blue and TCAA-CS reconstruction in red.

\paragraph{Nominal cycle}
The reconstruction closely matches the observed behavior across all signals, indicating that TCAA-CS captures normal cycle dynamics.

\paragraph{Load-related anomaly}
The opening phase is abnormally slow, with stalled motion visible in the position signal. The reconstruction predicts a normal trajectory, producing a large mismatch also reflected in current and voltage signals.

\paragraph{Kinematic anomaly}
The door reverses before reaching the fully open position. The reconstruction follows the expected trajectory, highlighting the missing motion.

\paragraph{Logical inconsistency}
The door-closed and door-locked signals activate while the position signal indicates ongoing motion. Each signal individually appears plausible, but their relationship is physically inconsistent. The reconstruction preserves nominal logical timing, exposing the inconsistency. This fault is primarily captured by $\mathcal{C}_{\text{cross}}$, illustrating the value of the cross-signal consistency term.

Overall, the qualitative analysis shows that the three scoring components complement each other, supporting a unified framework rather than separate handcrafted checks for each failure mode.

\section{Conclusion}
\label{sec:conclusion}

This paper presented TCAA-CS, a cycle-level unsupervised anomaly detection framework for railway passenger door monitoring. By combining dual-stream signal encoding, temporal attention pooling, and a triple hybrid anomaly score, the model detects both signal-level deviations and inter-signal inconsistencies across complete door operation cycles. Experiments on real industrial data show 93.8\% recall, 97.3\% precision, and a 0.5\% false-alarm rate, with all expert-confirmed anomalies detected. System-level evaluation on an NVIDIA Jetson AGX Xavier confirms real-time onboard deployment feasibility with millisecond-level latency and limited memory usage. Future work will focus on validation across larger multi-fleet datasets, additional door types, and adaptive consistency models that account for varying operating conditions while preserving low computational cost.

\section*{Acknowledgment}

The authors would like to thank Alstom for providing access to the operational railway door data used in this study. The authors also thank Eddy Doba and Nordine Saim from Alstom for their guidance and support throughout this work.

\bibliographystyle{IEEEtran}
\bibliography{references}

\end{document}